\documentclass[conference]{IEEEtran}

\makeatletter
\def\ps@IEEEtitlepagestyle{%
  \def\@oddfoot{\mycopyrightnotice}%
  \def\@oddhead{\hbox{}\@IEEEheaderstyle\leftmark\hfil\thepage}\relax
  \def\@evenhead{\@IEEEheaderstyle\thepage\hfil\leftmark\hbox{}}\relax
  \def\@evenfoot{}%
}
\def\mycopyrightnotice{%
  \begin{minipage}{\textwidth}
  \centering \scriptsize
  Copyright~\copyright~2022 IEEE.  Personal use of this material is permitted. Permission from IEEE must be obtained for all other uses, in any current or future media, including reprinting/republishing this material for advertising or promotional purposes, creating new collective works, for resale or redistribution to servers or lists, or reuse of any copyrighted component of this work in other works.
  \end{minipage}
}
\makeatother

\usepackage{cite}
\usepackage{amsmath,amssymb,amsfonts}
\usepackage{algorithmic}
\usepackage{graphicx}
\usepackage{textcomp}
\usepackage{xcolor}

\newcommand{\degree}{^{\circ}}
 \usepackage{multirow}
 
 \usepackage{array}
\newcommand{\PreserveBackslash}[1]{\let\temp=\\#1\let\\=\temp}
\newcolumntype{C}[1]{>{\PreserveBackslash\centering}m{#1}}
\newcolumntype{R}[1]{>{\PreserveBackslash\raggedleft}m{#1}}
\newcolumntype{L}[1]{>{\PreserveBackslash\raggedright}m{#1}}

\usepackage[bottom]{footmisc}
\usepackage{booktabs}

\usepackage[caption=false, subrefformat=parens, labelformat=parens]{subfig}
\usepackage{hyperref}
\hypersetup{
    colorlinks=true,
    allcolors=blue,
    pdftitle={HEATGait},
    }

\usepackage{amsmath}
\renewcommand\figureautorefname{Fig.}
\usepackage{balance}

\def\BibTeX{{\rm B\kern-.05em{\sc i\kern-.025em b}\kern-.08em
    T\kern-.1667em\lower.7ex\hbox{E}\kern-.125emX}}
\begin{document}

\title{HEATGait: Hop-Extracted Adjacency Technique in Graph Convolution based Gait Recognition}

\author{\IEEEauthorblockN{Md. Bakhtiar Hasan}
\IEEEauthorblockA{
\textit{Islamic University of Technology}\\
Gazipur, Bangladesh \\
bakhtiarhasan@iut-dhaka.edu}
\and
\IEEEauthorblockN{Tasnim Ahmed}
\IEEEauthorblockA{
\textit{Islamic University of Technology}\\
Gazipur, Bangladesh \\
tasnimahmed@iut-dhaka.edu}
\and
\IEEEauthorblockN{Md. Hasanul Kabir}
\IEEEauthorblockA{
\textit{Islamic University of Technology}\\
Gazipur, Bangladesh \\
hasanul@iut-dhaka.edu}
}

\maketitle

\begin{abstract}
Biometric authentication using gait has become a promising field due to its unobtrusive nature. Recent approaches in model-based gait recognition techniques utilize spatio-temporal graphs for the elegant extraction of gait features. However, existing methods often rely on multi-scale operators for extracting long-range relationships among joints resulting in biased weighting. In this paper, we present HEATGait, a gait recognition system that improves the existing multi-scale graph convolution by efficient hop-extraction technique to alleviate the issue. Combined with preprocessing and augmentation techniques, we propose a powerful feature extractor that utilizes ResGCN to achieve state-of-the-art performance in model-based gait recognition on the CASIA-B gait dataset.
\end{abstract}


\section{Introduction}
Gait recognition is one of the most popular biometric authentication techniques since it can be used without the cooperation of human subjects \cite{sepas2022deep}. In particular, model-based gait recognition focuses on extracting skeleton representation of human gait instead of using raw RGB videos. The representation consists of two-dimensional positions of key joints that are free of occlusions caused by carried items, clothing, etc., resulting in robust features facilitating gait recognition. It also utilizes less sensitive personal data.

Contemporary approaches on model-based gait recognition focus on extracting joint positions from gait video sequences leveraging the recent development in pose estimation techniques to generate a spatio-temporal graph considering the joints as vertices and the bones as edges \cite{liao2020model, teepe2021gaitgraph, wang2022multi, gao2022gaitd}. Since joints that are not connected via bones can have strong correlations, an optimal method for gait recognition should explore both local and global joint dependencies. Existing approaches accomplish this by applying graph convolutions using higher-order polynomials of the adjacency matrix representing the joint connections. This can enable the graph convolution to consider distant joints by capturing the length $n$ walk using the $n^\text{th}$ order polynomial. However, due to the presence of cyclic walks, this formulation results in higher edge weights for closer joints compared to the further ones. As a result, the effectiveness of this method in capturing information from distant joints is reduced affecting the overall performance of the system \cite{liu2020disentangling, sepas2022deep}.

In this paper, we present HEATGait, a gait recognition system that utilizes existing pose estimation architecture to extract and model human skeleton joints which are then further refined using preprocessing techniques. We employ a graph convolutional network equipped with an improved multi-scale feature aggregation technique utilizing hop-extraction to create adjacency matrix that mitigates the effect of biased weight. Our empirical results show state-of-the-art performance in model-based gait recognition.

\section{Related Works}
Model-based gait recognition considers human body structure modeled using joints and bones. The key benefit here is robustness to occlusion, viewing angle, and subject distance.

Earlier approaches of model-based gait recognition relied on extracting joint positions using external devices \cite{dikovski2014evaluation, ahmed2015dtw}. They extracted handcrafted features such as relative distance and angle between the joints in order to recognize gait using machine learning classifiers. However, due to severe feature engineering, these works are mostly concentrated on small datasets and often sensitive to environmental noises.

Recent approaches improved upon this in two ways. External devices to extract joint positions were phased out due to the recent advances in pose estimation techniques. Handcrafted features and machine learning-based classifiers were replaced with deep learning techniques. One of the earliest works in this regard utilized heatmaps generated from CNN to extract 2D features from each frame of gait sequence and fed them into an LSTM network to capture the temporal information embedded in the gait sequence \cite{feng2016learning}. In another work, a combination of CNN and LSTM was used to generate spatio-temporal features \cite{liao2017pose}. These works were further extended to include 3D gait features \cite{an2018improving, liao2020model}. However, these approaches depend on CNNs to extract gait features rather than utilizing pose information.

Considering the graph structure of gait features extracted using pose-estimation, graph convolutional networks (GCN) are being utilized in gait recognition \cite{sepas2022deep}. Li et al. pioneered the use of GCN in gait recognition along with Joints Relationship Pyramid Mapping (JRPM). They considered the joint positions as vertices and the bones connecting those joints as edges \cite{li2020model}. However, their best-performing model heavily depended on JRPM. Later, Teepe et al. proposed standalone GCN architecture utilizing ResGCN \cite{song2020stronger}, improving upon the performance of the earlier work. Later, Gao et al. introduced Gait-D architecture combining ST-GCN \cite{chen2018multi} and Canonical Polyadic Decomposition \cite{sorber2013optimization} to focus on relevant gait features to improve the recognition performance \cite{gao2022gaitd}. These methods combined the relative positions of the joints and their natural bone connection encoded using the adjacency matrix of the graph. Wang et al. extended this idea in multi-stream networks to integrate joint, bone and motion-based features \cite{wang2022multi}.

Even though the GCN-based approaches discussed above paved the way to enhance the performance of model-based gait recognition, their reliance on higher-order polynomials of the adjacency matrix for multi-scale feature aggregation makes them prone to biased weighting problem reducing the overall performance of the gait recognition system \cite{sepas2022deep}. This necessitates the introduction of a novel aggregation technique to capture the long-range relationship among joint positions while also avoiding the biased weighting problem.

\section{Proposed Approach}
\begin{figure}[b]
    \centering
    \includegraphics[width=0.95\columnwidth]{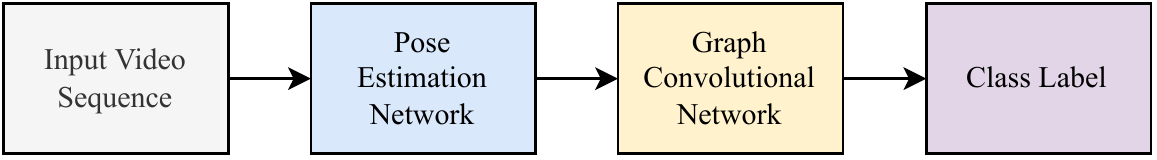}
    \caption{Overview of HEATGait Architecture}
    \label{fig:pipeline}
\end{figure}
The proposed gait recognition system takes gait video of a person as input. The videos are taken from the CASIA-B dataset \cite{yu2006framework}. At first, each frame of the input video is passed through a pose estimation network to estimate the joint positions representing human pose. The pose sequence is then preprocessed to remove predictions with low confidence. The modified sequence of poses is then fed to a graph convolutional network with residual connections to generate the class label corresponding to the predicted ID of the person. The overall pipeline of the proposed system is depicted in \figureautorefname~\ref{fig:pipeline}.

\subsection{CASIA-B Dataset}
CASIA-B is one of the most common datasets used to evaluate gait recognition architectures. It provides RGB images of the gait sequence of 126 human subjects. Each subject is captured from 11 viewing angles ($0\degree, 18\degree, \dots, 180\degree$) in three walking conditions (NM: normal walking, BH: walking while carrying a bag, CL: walking while wearing heavy clothes). For each subject 6 NM, 2 BG, and 2 CL sequences are provided for each of the 11 angles resulting in $(6 + 2 + 2) \times 11 = 110$ sequences per subject.

Since the authors of the dataset did not define any protocol to partition the data into training and test sets, we follow the split popularized by existing gait recognition literature to ensure fair comparison \cite{liao2020model, teepe2021gaitgraph, wang2022multi, gao2022gaitd}. The dataset was split into training, validation and test set maintaining the ratio of 48:12:40.

\subsection{Pose Extraction}
A gait video of a person can be considered as a sequence of frames $f_1, f_2, \dots, f_N$ where each frame is an RGB image and $N$ denotes the number of frames in the gait video. In order to perform graph convolution, the input video needs to be represented using graph, i.e., vertices and edges. To do that, each frame $f_i$ is passed through a pose estimation that extracts $M$ keypoints denoting the joint positions of the person. The network utilizes $M$ heatmaps, $H_1, H_2, \dots, H_M$ where $H_i$ denotes the position of the $i^\text{th}$ joint in 2D coordinates.

\begin{figure}[tb]
    \centering
    \includegraphics[width=0.8\columnwidth]{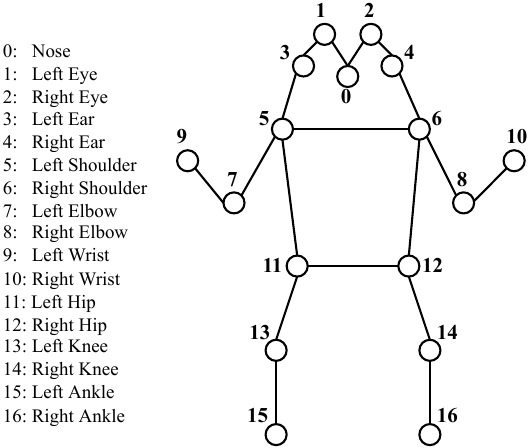}
    \caption{Joint positions extracted using HRNet. The bone configurations are adopted from \cite{teepe2021gaitgraph}}
    \label{fig:conf}
\end{figure}

In this work, we employ Higher Resolution Net (HRNet) \cite{sun2019deep}, pretrained on the COCO dataset \cite{lin2014microsoft} that is able to extract 17 joint positions from each of the frames. Since the bone information is not provided, we follow the configuration followed by \cite{teepe2021gaitgraph} (\figureautorefname~\ref{fig:conf}). Here, each joint is considered as a vertex and each bone is considered as an edge.

\begin{figure*}[tb]
    \captionsetup[subfloat]{format=hang, singlelinecheck=false}
    \centering
    \subfloat[Biased Weighting Problem: Vertices that are close to the central vertex get higher weights from higher-order polynomial of adjacency matrix, hampering the effectiveness of the long-range relationship.\label{fig:biased1}]{
    \includegraphics[width=0.45\textwidth]{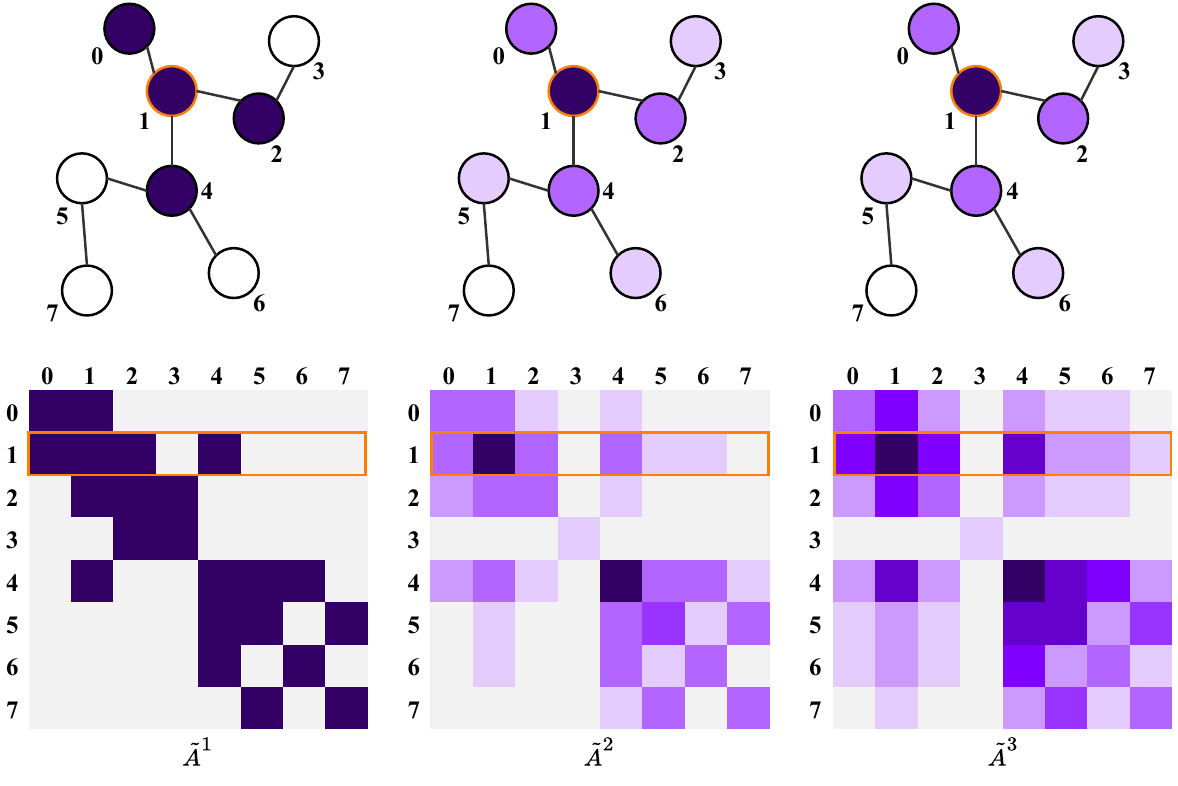}	
    }
    \hspace{0.5cm}\subfloat[Proposed Solution: Hop-extracted adjacency matrix ensures equal contribution for each neighbor in a certain distance while also keeping the identify features intact.\label{fig:biased2}]{
    \includegraphics[width=0.45\textwidth]{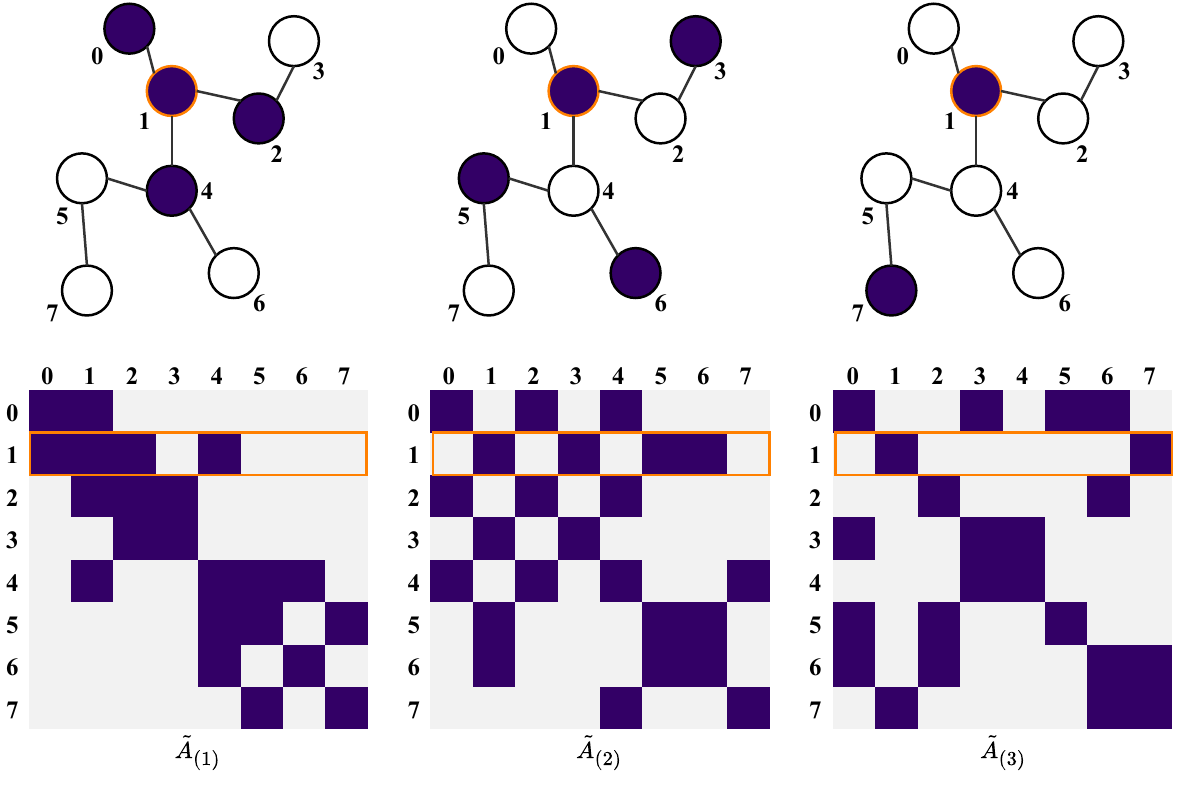}	
    }
    \caption{Demonstration of biased weighting problem and the proposed solution on a simple graph. Lighter color denotes lower weight and vice-versa. Self-loops are not shown to ensure the clarity of the image. Here, 1 is considered as the center vertex.}
    \label{fig:biased}
\end{figure*}

\subsection{Preprocessing}
\subsubsection{Removal of Low Confidence Frames}
In addition to providing the 2D coordinate of the 17 joints of a frame, HRNet generates a score for each joint denoting the confidence of the predicted joint position. The higher the score, the better the prediction. Since the relative position of the joints in subsequent frames is an important feature to recognize the gait, we hypothesize that a joint prediction with low confidence score can hamper the recognition performance. Particularly, in BG and CL walking conditions where the person is occluded by bag and clothes, respectively, the confidence of the HRNet in predicting joint positions can be significantly low. To counter this effect, we calculate the average confidence for each video frame:
\begin{equation}
    \text{Confidence} = \frac{\sum\limits_{i=1}^{M}c_i}{M}
\end{equation}
where $c_i$ denotes the confidence of the $i^\text{th}$ joint. We remove a video frame if the average confidence is below $60\%$.

\subsubsection{Data Augmentation}
To ensure the robustness of the proposed system, data augmentation techniques are used. First, we simulate the effect of a person walking backwards by inverting the order of the video frames. Again, the extracted joint positions were mirrored based on the center of gravity of the graph along the vertical axis to imitate the same person walking in the reverse direction. Additionally, small Gaussian noise was added to each of the joints to mimic the inaccuracies introduced by pose estimation techniques. All these augmentations were performed during the training phase of the network.

\subsection{Graph Convolution}
Let, $\mathcal{G} = (\mathcal{V}, \mathcal{E})$ be the human skeleton graph. Here, $\mathcal{V} = \{v_1, v_2, \dots, v_M\}$ is the set of $M$ vertices each corresponding to a joint, and $\mathcal{E}$ is the set of edges corresponding to the bones. Each vertex contains a pair of values corresponding to the x and y-coordinates of the joint. The edge information is encoded using an adjacency matrix $\mathbf{A} \in \mathbb{R}^{M \times M}$ where
\begin{equation}
    A_{i, j} = \begin{cases}
    1, & \text{there exists an edge between }v_i\text{ and }v_j\\
    0, & \text{otherwise}
    \end{cases}
\end{equation}
Note that, since $\mathcal{G}$ is undirected, $\mathbf{A}$ is symmetric.

The human gait is represented using a vertex features set $\mathbf{X} = \{x_{i, j} \in \mathbb{R}^C \, | \, i, j \in \mathbb{Z}; 1 \leq i \leq N; 1 \leq j \leq M\}$ where $x_{i, j}$ is the $C$-dimensional feature vector for vertex $v_j$ at frame $i$. Here, $\mathbf{X} \in \mathbb{R}^{N\times M\times C}$. Since we are using 2D coordinates of the joints, we have $C = 2$.

That means, $\mathbf{A}$ and $\mathbf{X}$ can represent the structural and feature information of a gait sequence, respectively.

Now, the layer-wise update for GCNs applied for frame $f$ is defined as:
\begin{equation}
    X_f^{(l+1)} = \sigma\left(\Tilde{\mathbf{D}}^{\text{-}\frac{1}{2}}\Tilde{\mathbf{A}}\Tilde{\mathbf{D}}^{\text{-}\frac{1}{2}}\mathbf{X}_f^{(l)}\Theta^{(l)}\right)
\end{equation}

where $\Tilde{\mathbf{A}} = \mathbf{A} + \mathbf{I}$ denotes the conservation of identity features by the introduction of self-loops, $\Tilde{\mathbf{D}}$ is a matrix containing the degrees of $\Tilde{\mathbf{A}}$ in its main diagonal, $\Theta^{(l)}$ denotes the learnable weight matrix for layer $l$, and $\sigma(\cdot)$ denotes the activation function. The diagonal degree matrix is used to normalize the features in order to prevent vanishing/exploding gradients \cite{kipf2017semi}.
\subsection{Addressing Biased Weighting Problem}

To aggregate multi-scale structural information, the spatial aggregation framework incorporates higher-order polynomials of the adjacency matrix:
\begin{equation}
    X_f^{(l+1)} = \sigma\left(\sum_{k=0}^K \hat{\mathbf{A}}^k\mathbf{X}_f^{(l)}\Theta^{(l)}_{(k)}\right) \label{eq:orig}
\end{equation}

where $K$ denotes the scale of the aggregation, $\hat{A} = \Tilde{\mathbf{D}}^{\text{-}\frac{1}{2}}\Tilde{\mathbf{A}}\Tilde{\mathbf{D}}^{\text{-}\frac{1}{2}}$. Note that, $A^k_{i, j} = A^k_{j, i}$ is the total number of length $k$ walks between $v_i$ and $v_j$. That means, $\hat{\mathbf{A}}^k\mathbf{X}_f^{(l)}$ can be used to perform feature aggregation weighted by the number of such walks.

Unfortunately, since walks consist of hops between vertex $i$ to vertex $j$, where $i$ can be equal to $j$, there can be cyclic walks concentrated in the originating vertex. Additionally, the existence of self-loops to preserve identity features can further increase such walks. As a result, the adjacency matrix contains higher values for the vertices close to the originating vertex and lower values for that of further (\figureautorefname~\ref{fig:biased}\subref{fig:biased1}). This creates a bias in feature aggregation, rendering the process less effective in capturing long-range relationship between joints.

\begin{figure*}[tb]
    \centering
    \includegraphics[width=0.7\textwidth]{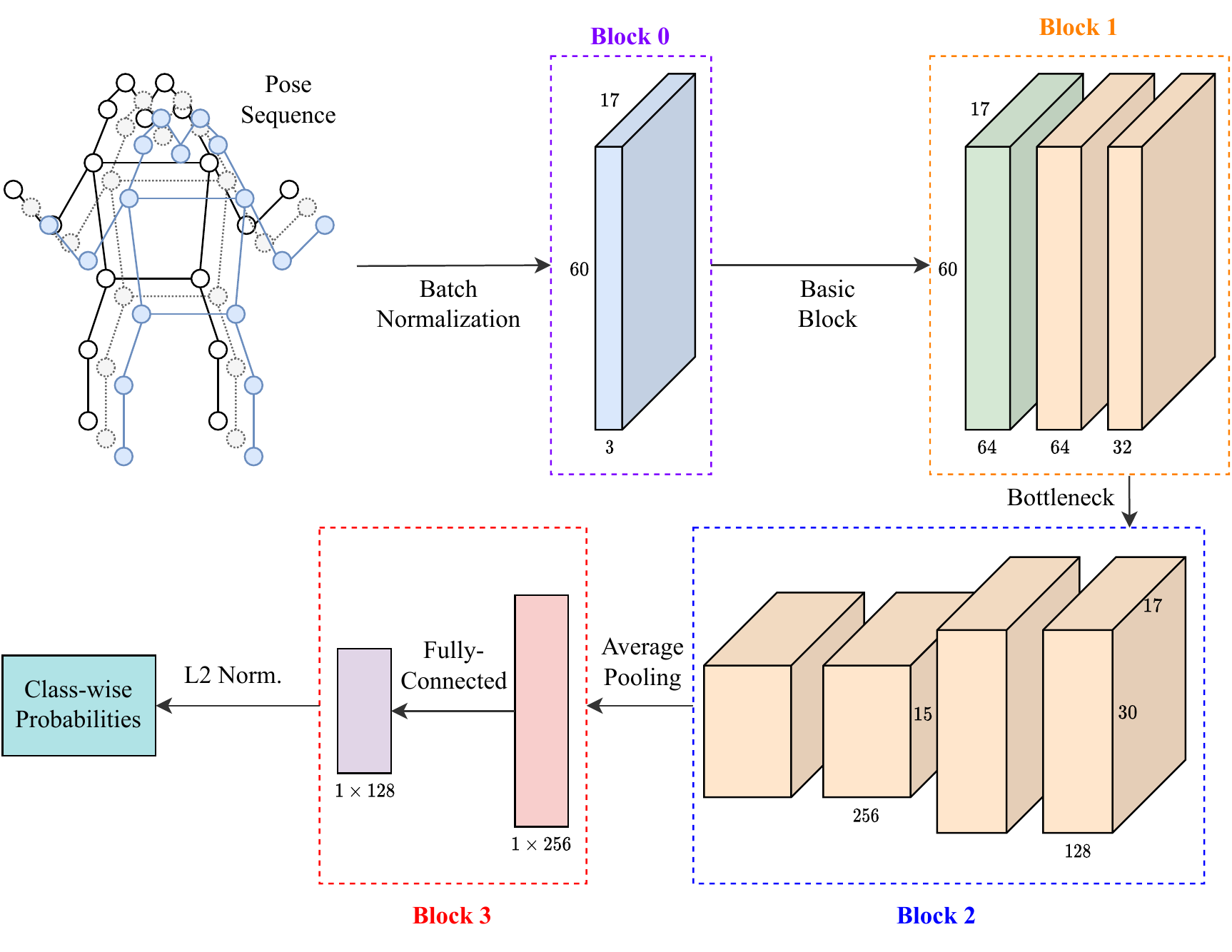}
    \caption{Overview of the ResGCN architecture for extraction of 17 joints from a video containing 60 frames. Here, each cube represents the feature maps (not to scale) after going through different layers. The blue cube is the output of batch normalization layer, green cube is the output of basic block, yellow cube is the output of bottleneck block.}
    \label{fig:resgcn}
\end{figure*}

We address the above issue by defining a $k$-adjacency matrix as:
\begin{equation}
     \left[\Tilde{\mathbf{A}}_{(k)}\right]_{i, j} = \begin{cases}
    1, & \text{if } d(v_i, v_j) = k\\
    1, & \text{if } i = j\\
    0, & \text{otherwise}
    \end{cases}\label{eq:form}
\end{equation}

where $d(v_i, v_j)$ is the shortest distance between $v_i$ and $v_j$ considering the number of hops (\figureautorefname~\ref{fig:biased}\subref{fig:biased2}). Since $\mathcal{G}$ is undirected, the value is calculated using Breadth-First Search (BFS) algorithm to find the $k$-hop neighbors. Note that, $\Tilde{\mathbf{A}}_{(1)} = \Tilde{\mathbf{A}}$ and $\Tilde{\mathbf{A}}_{(1)} = \mathbf{I}$.

Thereafter, incorporating \equationautorefname~\ref{eq:form} with \equationautorefname~\ref{eq:orig}, we get:
\begin{equation}
    X_f^{(l+1)} = \sigma\left(\sum_{k=0}^K \Tilde{\mathbf{D}}_{(k)}^{\text{-}\frac{1}{2}}\Tilde{\mathbf{A}}_{(k)}\Tilde{\mathbf{D}}_{(k)}^{\text{-}\frac{1}{2}}\mathbf{X}_f^{(l)}\Theta^{(l)}_{(k)}\right)\label{eq:final}
\end{equation}

In contrast with \equationautorefname~\ref{eq:orig} where the total number of length $k$ walks is dependent on length $k - 1$ walks, \equationautorefname~\ref{eq:final} provides equal importance to the vertices in closer and further neighborhoods alleviating the biased weighting problem. Consequently, this results in effective consideration of long-range relationships.

\begin{table*}[tb]
    \centering
    \caption{Performance comparison with the state-of-the-art models for model-based gait recognition.}
    \label{tab:sota}
    \begin{tabular}{c | C{2.4cm} | c c c c c c c c c c c | c}
    \toprule
        \multicolumn{2}{l |}{Gallery NM\#1-4} & \multicolumn{11}{c |}{Viewing angles} & \multirow{2}{*}{Mean}\\
        Probe & Ref. & $0\degree$ & $18\degree$ & $36\degree$ & $54\degree$ & $72\degree$ & $90\degree$ & $108\degree$ & $126\degree$ & $144\degree$ & $162\degree$ & $180\degree$ & \\\midrule
        \multirow{8}{*}{NM\#5-6} & PTSN \cite{liao2017pose} & $34.5$ & $45.6$ & $49.6$ & $51.3$ & $52.7$ & $52.3$ & $53$ & $50.8$ & $52.2$ & $48.3$ & $31.4$ & $47.4$\\
        & PTSN-3D \cite{an2018improving} & $38.7$ & $50.2$ & $55.9$ & $56$ & $56.7$ & $54.6$ & $54.8$ & $56$ & $54.1$ & $52.4$ & $40.2$ & $51.8$\\
        & PoseGait \cite{liao2020model} & $55.3$ & $69.6$ & $73.9$ & $75$ & $68$ & $68.2$ & $71.1$ & $72.9$ & $76.1$ & $70.4$ & $55.4$ & $68.7$\\
        & JointsGait \cite{li2020model} & $68.1$ & $73.6$ & $77.9$ & $76.4$ & $77.5$ & $79.1$ & $78.4$ & $76$ & $69.5$ & $71.9$ & $70.1$ & $74.4$\\
        & GaitGraph \cite{teepe2021gaitgraph} & $85.3$ & $88.5$ & $91$ & $92.5$ & $87$ & $86.5$ & $88.4$ & $89.2$ & $87.9$ & $85.9$ & $81.9$ & $87.6$\\
        & Gait-D\cite{gao2022gaitd} & $87.7$ & $92.5$ & $93.6$ & $\mathbf{95.7}$ & $\mathbf{93.3}$ & $92.4$ & $92.8$ & $93.4$ & $90.6$ & $88.6$ & $87.3$ & $91.6$\\
        & MS-Gait \cite{wang2022multi} & $89.4$ & $91.7$ & $91.6$ & $90.2$ & $90.6$ & $90.6$ & $90.4$ & $90.9$ & $90.4$ & $88.5$ & $85.6$ & $90.0$\\
        & HEATGait (Ours) & $\mathbf{91.7}$ & $\mathbf{93.8}$ & $\mathbf{93.8}$ & $94.7$ & $92.6$ & $\mathbf{94.6}$ & $\mathbf{94.3}$ & $\mathbf{94.4}$ & $\mathbf{93.2}$ & $\mathbf{91.5}$ & $\mathbf{91.5}$ & $\mathbf{93.3}$\\\midrule
        \multirow{8}{*}{BG\#1-2} & PTSN \cite{liao2017pose} & $22.4$ & $29.8$ & $29.6$ & $29.2$ & $32.5$ & $31.5$ & $32.1$ & $31$ & $27.3$ & $28.1$ & $18.2$ & $28.3$\\
        & PTSN-3D \cite{an2018improving} & $27.7$ & $32.7$ & $37.4$ & $35$ & $37.1$ & $37.5$ & $37.7$ & $36.9$ & $33.8$ & $31.8$ & $27$ & $34.1$\\
        & PoseGait \cite{liao2020model} & $35.3$ & $47.2$ & $52.4$ & $46.9$ & $45.5$ & $43.9$ & $46.1$ & $48.1$ & $49.4$ & $43.6$ & $31.1$ & $44.5$\\
        & JointsGait \cite{li2020model} & $54.3$ & $59.1$ & $60.6$ & $59.7$ & $63$ & $65.7$ & $62.4$ & $59$ & $58.1$ & $58.6$ & $50.1$ & $59.1$\\
        & GaitGraph \cite{teepe2021gaitgraph} & $75.8$ & $76.7$ & $75.9$ & $76.1$ & $71.4$ & $73.9$ & $78$ & $74.7$ & $75.4$ & $75.4$ & $69.2$ & $74.8$\\
        & Gait-D \cite{gao2022gaitd} & $78.2$ & $80.1$ & $79.3$ & $80.2$ & $78.4$ & $77.6$ & $80.4$ & $78.6$ & $79.1$ & $80.2$ & $76.5$ & $79.0$\\
        & MS-Gait \cite{wang2022multi} & $75.7$ & $84.8$ & $83.7$ & $83.2$ & $80.6$ & $80.1$ & $82.2$ & $79.8$ & $79.1$ & $75.9$ & $71.1$ & $79.7$\\
        & HEATGait (Ours) & $\mathbf{86.9}$ & $\mathbf{87.2}$ & $\mathbf{89.6}$ & $\mathbf{89.5}$ & $\mathbf{86.5}$ & $\mathbf{88}$ & $\mathbf{89.4}$ & $\mathbf{86.9}$ & $\mathbf{87.1}$ & $\mathbf{85.6}$ & $\mathbf{85.5}$ & $\mathbf{87.5}$\\\midrule
        \multirow{8}{*}{CL\#1-2} & PTSN \cite{liao2017pose} & $14.2$ & $17.1$ & $17.6$ & $19.3$ & $19.5$ & $20$ & $20.1$ & $17.3$ & $16.5$ & $18.1$ & $14$ & $17.6$\\
        & PTSN-3D \cite{an2018improving} & $15.8$ & $17.2$ & $19.9$ & $20$ & $22.3$ & $24.3$ & $28.1$ & $23.8$ & $20.9$ & $23$ & $17$ & $21.1$\\
        & PoseGait \cite{liao2020model} & $24.3$ & $29.7$ & $41.3$ & $38.8$ & $38.2$ & $38.5$ & $41.6$ & $44.9$ & $42.2$ & $33.4$ & $22.5$ & $35.9$\\
        & JointsGait \cite{li2020model} & $48.1$ & $46.9$ & $49.6$ & $50.5$ & $51$ & $52.3$ & $49$ & $46$ & $48.7$ & $53.6$ & $52$ & $49.8$\\
        & GaitGraph \cite{teepe2021gaitgraph} & $69.6$ & $66.1$ & $68.8$ & $67.2$ & $64.5$ & $62$ & $69.5$ & $65.6$ & $65.7$ & $66.1$ & $64.3$ & $66.3$\\
        & Gait-D \cite{gao2022gaitd} & $73.2$ & $71.7$ & $75.4$ & $73.2$ & $74.6$ & $72.3$ & $74.1$ & $70.5$ & $69.4$ & $71.2$ & $66.7$ & $72.0$\\
        & MS-Gait \cite{wang2022multi} & $75.1$ & $79.7$ & $80.5$ & $\mathbf{84.7}$ & $\mathbf{84}$ & $82.4$ & $79.8$ & $80.4$ & $78.3$ & $78$ & $70.9$ & $79.4$\\
        & HEATGait (Ours) & $\mathbf{82.8}$ & $\mathbf{81.3}$ & $\mathbf{87.4}$ & $83.9$ & $82.5$ & $\mathbf{84.1}$ & $\mathbf{83.9}$ & $\mathbf{81.1}$ & $\mathbf{80.7}$ & $\mathbf{78.5}$ & $\mathbf{78.8}$ & $\mathbf{82.3}$\\\bottomrule
    \end{tabular}
\end{table*}

\subsection{Model Architecture}
We have adapted the ResGCN architecture that have been used for action recognition to suit our task. The architecture, as shown in \figureautorefname~\ref{fig:resgcn}, consists of a set of ResGCN blocks composed of Graph Convolution to extract spatial features, 2D Convolution to extract temporal features, and a set of residual connections. Some of the internal layers may contain a set of bottleneck structures. The output of the internal layers are passed through an average pooling and a fully-connected layer sequentially to map the extracted features with the class labels.

The first two blocks, Block 0 and Block 1 work as the input branch. Output from the Block 1 is fed into the next main stream block that produces the input for the final block which, in turn, produces the class-wise probabilities by normalizing the outputs.

The initial batch normalized input is given as an input to the subsequent spatial and temporal `Basic' blocks. The `Basic' block is a sequential set of convolution and batch normalization layers that conditionally has a residual connection. The final output of the `Basic' block is produced by passing the obtained output through a ReLU activation function.

The `Bottleneck' blocks are used to reconstruct the input using a sequence of up-convolution and down-convolution where each convolution layer is followed by a batch normalization layer. In case of the block being residual, a convoluted and batch normalized form of the input is added with the output. Final output is produced by passing the output through ReLU activation function.

Afterwards, the features are extracted smoothly instead of retaining more pronounced features like edges by applying an average pooling on the output feature map from the last bottleneck block of Block 2. Finally, the feature vector is mapped to the output units through a fully-connected layer.

\section{Results and Discussion}
In this section, we compare our proposed approach with other state-of-the-art techniques in public gait dataset CASIA-B. The comparison is performed considering different views and different walking conditions with other model-based approaches.

\subsection{Experimental Setup}
The proposed system was trained in a Python environment in Google Colab \footnote{\url{https://colab.research.google.com/}}. Considering the noise introduced due to the approximation of joint position, we utilize Adam optimizer \cite{diederik2015adam}. The architecture was trained in multiple cycles. Each cycle consisted of 100 epochs. The initial learning rate was set to $0.01$. After each cycle, it was reduced to $10\%$ of the previous cycle down to a minimum of $10^{-5}$. This was done to ensure that the model does not overfit to the training samples. Supervised contrastive loss \cite{khosla2020supervised} was used to calculate the loss.

\subsection{Comparison with State-of-the-Art Methods}
\tableautorefname~\ref{tab:sota} presents a comparison of our proposed system with the state-of-the-art models in gait recognition. Our system achieves a commendable performance throughout different viewing angles and walking conditions. 

We implemented the architecture of \cite{teepe2021gaitgraph} as our baseline and performed modifications upon it for the improvement of recognition performance. According to \tableautorefname~\ref{tab:sota}, a significant improvement can be noticed compared to the baseline. This improvement can be attributed to our conservative training process which was undertaken to ensure that the model does not overfit the training samples, a problem that is persistent in \cite{teepe2021gaitgraph}. 

In all walking conditions, our system outperforms the existing state-of-the-art for all but two viewing angles. Even in the case of $54\degree$ and $72\degree$ viewing angles, the performance of our system is comparable. All in all, our proposed system achieves state-of-the-art performance compared to the existing literature.

\subsection{Ablation Study}
Compared to the baseline shown in \tableautorefname~\ref{tab:ablation}, our preprocessing techniques were able to improve the performance in all aspects. Specifically, considering the BG and CL walking conditions are more prone to inaccuracies in pose estimation technique, the improvement of performance in these conditions are more prominent compared to that of NM as hypothesized. In this regard, it is worth-mentioning that the preprocessing removes $0.9\%$ frames from the dataset. Reduction of noisy and inaccurate frames helped the model converge faster and avoid overfitting. This resulted in a $2.33\%$ increase of accuracy on average.

\begin{table}[tb]
    \centering
    \caption{Ablation study on the CASIA-B Dataset}
    \label{tab:ablation}
    \begin{tabular}{C{3cm} c c c}
        \toprule
        \multirow{2}{*}{Strategy} & \multicolumn{3}{c}{Accuracy (\%)}\\
         & NM & BG & CL\\
        \midrule
        Baseline (ResGCN) & $87.5$ & $75.3$ & $67.1$\\
        Baseline + Preprocessing & $88.9$ & $78.2$ & $69.8$\\
        Baseline + Preprocessing + Hop Extraction & $93.3$ & $87.5$ & $82.3$\\
        \bottomrule
    \end{tabular}
\end{table}

Combining the hop extraction technique with the baseline and using preprocessed dataset increases the accuracy by $11.07\%$ on average compared to the baseline proving the effectiveness of our approach.

\section{Conclusion}
In this paper, we present a novel hop-extracted adjacency technique to address the biased weighting problem persistent in existing model-based gait recognition. In this regard, we propose a gait recognition system leveraging efficient preprocessing techniques and recent advancement in pose estimation and graph convolution approaches. Our empirical results show that our system outperforms the existing architectures by a sizable margin.

Most of the existing works on gait recognition either focus on appearance-based or model-based features. Future research efforts can be concentrated on combining these two features via ensemble technique to further improve the performance. On the other hand, the existing datasets on gait recognition provide videos captured on laboratory settings, which might be essential considering the requirements to capture the same gait from different viewing angles and poses. However, videos captured in natural settings can provide much more realistic environments and help researchers experiment how their systems perform in real-life scenarios.

\section*{Acknowledgment}

The authors are thankful to all the faculty members of Department of Computer Science and Engineering, Islamic University of Technology for their continuous support and suggestions throughout the entire study.
\balance
\bibliographystyle{IEEEtran}
\bibliography{ref}

\begin{thebibliography}{10}
\providecommand{\url}[1]{#1}
\csname url@samestyle\endcsname
\providecommand{\newblock}{\relax}
\providecommand{\bibinfo}[2]{#2}
\providecommand{\BIBentrySTDinterwordspacing}{\spaceskip=0pt\relax}
\providecommand{\BIBentryALTinterwordstretchfactor}{4}
\providecommand{\BIBentryALTinterwordspacing}{\spaceskip=\fontdimen2\font plus
\BIBentryALTinterwordstretchfactor\fontdimen3\font minus
  \fontdimen4\font\relax}
\providecommand{\BIBforeignlanguage}[2]{{%
\expandafter\ifx\csname l@#1\endcsname\relax
\typeout{** WARNING: IEEEtran.bst: No hyphenation pattern has been}%
\typeout{** loaded for the language `#1'. Using the pattern for}%
\typeout{** the default language instead.}%
\else
\language=\csname l@#1\endcsname
\fi
#2}}
\providecommand{\BIBdecl}{\relax}
\BIBdecl

\bibitem{sepas2022deep}
\BIBentryALTinterwordspacing
A.~Sepas-Moghaddam and A.~Etemad, ``{Deep Gait Recognition: A Survey},''
  \emph{IEEE Transactions on Pattern Analysis and Machine Intelligence}, pp.
  1--1, 2022. [Online]. Available:
  \url{https://ieeexplore.ieee.org/abstract/document/9714177}
\BIBentrySTDinterwordspacing

\bibitem{liao2020model}
\BIBentryALTinterwordspacing
R.~Liao, S.~Yu, W.~An, and Y.~Huang, ``{A model-based gait recognition method
  with body pose and human prior knowledge},'' \emph{Pattern Recognition},
  vol.~98, p. 107069, 2020. [Online]. Available:
  \url{https://www.sciencedirect.com/science/article/pii/S003132031930370X}
\BIBentrySTDinterwordspacing

\bibitem{teepe2021gaitgraph}
\BIBentryALTinterwordspacing
T.~Teepe, A.~Khan, J.~Gilg, F.~Herzog, S.~Hörmann, and G.~Rigoll,
  ``{Gaitgraph: Graph Convolutional Network for Skeleton-Based Gait
  Recognition},'' in \emph{2021 IEEE International Conference on Image
  Processing (ICIP)}.\hskip 1em plus 0.5em minus 0.4em\relax Anchorage, Alaska,
  USA: {IEEE}, 2021, pp. 2314--2318. [Online]. Available:
  \url{https://ieeexplore.ieee.org/abstract/document/9506717}
\BIBentrySTDinterwordspacing

\bibitem{wang2022multi}
\BIBentryALTinterwordspacing
L.~Wang, J.~Chen, Z.~Chen, Y.~Liu, and H.~Yang, ``{Multi-stream part-fused
  graph convolutional networks for skeleton-based gait recognition},''
  \emph{Connection Science}, vol.~34, no.~1, pp. 652--669, 2022. [Online].
  Available:
  \url{https://www.tandfonline.com/doi/full/10.1080/09540091.2022.2026294}
\BIBentrySTDinterwordspacing

\bibitem{gao2022gaitd}
\BIBentryALTinterwordspacing
S.~Gao, J.~Yun, Y.~Zhao, and L.~Liu, ``{Gait-D: Skeleton-based gait feature
  decomposition for gait recognition},'' \emph{IET Computer Vision}, vol.~16,
  no.~2, pp. 111--125, 2022. [Online]. Available:
  \url{https://ietresearch.onlinelibrary.wiley.com/doi/abs/10.1049/cvi2.12070}
\BIBentrySTDinterwordspacing

\bibitem{liu2020disentangling}
\BIBentryALTinterwordspacing
Z.~Liu, H.~Zhang, Z.~Chen, Z.~Wang, and W.~Ouyang, ``{Disentangling and
  Unifying Graph Convolutions for Skeleton-Based Action Recognition},'' in
  \emph{2020 IEEE/CVF Conference on Computer Vision and Pattern Recognition
  (CVPR)}.\hskip 1em plus 0.5em minus 0.4em\relax {IEEE}, 2020, pp. 140--149.
  [Online]. Available:
  \url{https://openaccess.thecvf.com/content_CVPR_2020/html/Liu_Disentangling_and_Unifying_Graph_Convolutions_for_Skeleton-Based_Action_Recognition_CVPR_2020_paper.html}
\BIBentrySTDinterwordspacing

\bibitem{dikovski2014evaluation}
\BIBentryALTinterwordspacing
B.~Dikovski, G.~Madjarov, and D.~Gjorgjevikj, ``{Evaluation of different
  feature sets for gait recognition using skeletal data from Kinect},'' in
  \emph{2014 37th International Convention on Information and Communication
  Technology, Electronics and Microelectronics (MIPRO)}.\hskip 1em plus 0.5em
  minus 0.4em\relax Opatija, Croatia: {IEEE}, 2014, pp. 1304--1308. [Online].
  Available: \url{https://ieeexplore.ieee.org/abstract/document/6859769}
\BIBentrySTDinterwordspacing

\bibitem{ahmed2015dtw}
\BIBentryALTinterwordspacing
F.~Ahmed, P.~P. Paul, and M.~L. Gavrilova, ``{DTW-based kernel and rank-level
  fusion for 3D gait recognition using Kinect},'' \emph{The visual computer},
  vol.~31, no.~6, pp. 915--924, 2015. [Online]. Available:
  \url{https://link.springer.com/article/10.1007/s00371-015-1092-0}
\BIBentrySTDinterwordspacing

\bibitem{feng2016learning}
\BIBentryALTinterwordspacing
Y.~Feng, Y.~Li, and J.~Luo, ``{Learning effective Gait features using LSTM},''
  in \emph{2016 23rd International Conference on Pattern Recognition
  (ICPR)}.\hskip 1em plus 0.5em minus 0.4em\relax Cancun, Mexico: {IEEE}, 2016,
  pp. 325--330. [Online]. Available:
  \url{https://ieeexplore.ieee.org/abstract/document/7899654}
\BIBentrySTDinterwordspacing

\bibitem{liao2017pose}
\BIBentryALTinterwordspacing
R.~Liao, C.~Cao, E.~B. Garcia, S.~Yu, and Y.~Huang, ``{Pose-Based
  Temporal-Spatial Network (PTSN) for Gait Recognition with Carrying and
  Clothing Variations},'' in \emph{Biometric Recognition}, J.~Zhou, Y.~Wang,
  Z.~Sun, Y.~Xu, L.~Shen, J.~Feng, S.~Shan, Y.~Qiao, Z.~Guo, and S.~Yu,
  Eds.\hskip 1em plus 0.5em minus 0.4em\relax Cham: {Springer International
  Publishing}, 2017, pp. 474--483. [Online]. Available:
  \url{https://link.springer.com/chapter/10.1007/978-3-319-69923-3_51}
\BIBentrySTDinterwordspacing

\bibitem{an2018improving}
\BIBentryALTinterwordspacing
W.~An, R.~Liao, S.~Yu, Y.~Huang, and P.~C. Yuen, ``{Improving Gait Recognition
  with 3D Pose Estimation},'' in \emph{Biometric Recognition}, J.~Zhou,
  Y.~Wang, Z.~Sun, Z.~Jia, J.~Feng, S.~Shan, K.~Ubul, and Z.~Guo, Eds.\hskip
  1em plus 0.5em minus 0.4em\relax Cham: {Springer International Publishing},
  2018, pp. 137--147. [Online]. Available:
  \url{https://link.springer.com/chapter/10.1007/978-3-319-97909-0_15}
\BIBentrySTDinterwordspacing

\bibitem{li2020model}
\BIBentryALTinterwordspacing
N.~Li, X.~Zhao, and C.~Ma, ``{A model-based Gait Recognition Method based on
  Gait Graph Convolutional Networks and Joints Relationship Pyramid Mapping},''
  \emph{CoRR}, vol. abs/2005.08625, 2020. [Online]. Available:
  \url{https://arxiv.org/abs/2005.08625}
\BIBentrySTDinterwordspacing

\bibitem{song2020stronger}
\BIBentryALTinterwordspacing
Y.-F. Song, Z.~Zhang, C.~Shan, and L.~Wang, \emph{{Stronger, Faster and More
  Explainable: A Graph Convolutional Baseline for Skeleton-Based Action
  Recognition}}.\hskip 1em plus 0.5em minus 0.4em\relax New York, NY, USA:
  {Association for Computing Machinery}, 2020, p. 1625–1633. [Online].
  Available: \url{https://dl.acm.org/doi/abs/10.1145/3394171.3413802}
\BIBentrySTDinterwordspacing

\bibitem{chen2018multi}
\BIBentryALTinterwordspacing
Z.~Chen, S.~Li, B.~Yang, Q.~Li, and H.~Liu, ``{Multi-Scale Spatial Temporal
  Graph Convolutional Network for Skeleton-Based Action Recognition},''
  \emph{Proceedings of the AAAI Conference on Artificial Intelligence},
  vol.~35, no.~2, pp. 1113--1122, May 2021. [Online]. Available:
  \url{https://ojs.aaai.org/index.php/AAAI/article/view/16197}
\BIBentrySTDinterwordspacing

\bibitem{sorber2013optimization}
\BIBentryALTinterwordspacing
L.~Sorber, M.~Van~Barel, and L.~De~Lathauwer, ``{Optimization-Based Algorithms
  for Tensor Decompositions: Canonical Polyadic Decomposition, Decomposition in
  Rank-\$(L\_r,L\_r,1)\$ Terms, and a New Generalization},'' \emph{SIAM Journal
  on Optimization}, vol.~23, no.~2, pp. 695--720, 2013. [Online]. Available:
  \url{https://epubs.siam.org/doi/abs/10.1137/120868323}
\BIBentrySTDinterwordspacing

\bibitem{yu2006framework}
\BIBentryALTinterwordspacing
S.~Yu, D.~Tan, and T.~Tan, ``A framework for evaluating the effect of view
  angle, clothing and carrying condition on gait recognition,'' in \emph{18th
  International Conference on Pattern Recognition (ICPR'06)}, vol.~4.\hskip 1em
  plus 0.5em minus 0.4em\relax {IEEE}, 2006, pp. 441--444. [Online]. Available:
  \url{https://ieeexplore.ieee.org/abstract/document/1699873}
\BIBentrySTDinterwordspacing

\bibitem{sun2019deep}
\BIBentryALTinterwordspacing
K.~Sun, B.~Xiao, D.~Liu, and J.~Wang, ``Deep high-resolution representation
  learning for human pose estimation,'' in \emph{2019 IEEE/CVF Conference on
  Computer Vision and Pattern Recognition (CVPR)}.\hskip 1em plus 0.5em minus
  0.4em\relax Long Beach, California: {IEEE}, 2019, pp. 5686--5696. [Online].
  Available: \url{https://ieeexplore.ieee.org/document/8953615}
\BIBentrySTDinterwordspacing

\bibitem{lin2014microsoft}
\BIBentryALTinterwordspacing
T.-Y. Lin, M.~Maire, S.~Belongie, J.~Hays, P.~Perona, D.~Ramanan,
  P.~Doll{\'a}r, and C.~L. Zitnick, ``Microsoft coco: Common objects in
  context,'' in \emph{Computer Vision -- ECCV 2014}, D.~Fleet, T.~Pajdla,
  B.~Schiele, and T.~Tuytelaars, Eds.\hskip 1em plus 0.5em minus 0.4em\relax
  Cham: Springer International Publishing, 2014, pp. 740--755. [Online].
  Available:
  \url{https://link.springer.com/chapter/10.1007/978-3-319-10602-1_48}
\BIBentrySTDinterwordspacing

\bibitem{kipf2017semi}
\BIBentryALTinterwordspacing
T.~N. Kipf and M.~Welling, ``Semi-supervised classification with graph
  convolutional networks,'' in \emph{5th International Conference on Learning
  Representations, {ICLR} 2017, Toulon, France, April 24-26, 2017, Conference
  Track Proceedings}.\hskip 1em plus 0.5em minus 0.4em\relax OpenReview.net,
  2017. [Online]. Available: \url{https://openreview.net/forum?id=SJU4ayYgl}
\BIBentrySTDinterwordspacing

\bibitem{diederik2015adam}
\BIBentryALTinterwordspacing
D.~P. Kingma and J.~Ba, ``{Adam: A Method for Stochastic Optimization},'' in
  \emph{3rd International Conference on Learning Representations (ICLR)},
  Y.~Bengio and Y.~LeCun, Eds., 2015. [Online]. Available:
  \url{http://arxiv.org/abs/1412.6980}
\BIBentrySTDinterwordspacing

\bibitem{khosla2020supervised}
\BIBentryALTinterwordspacing
P.~Khosla, P.~Teterwak, C.~Wang, A.~Sarna, Y.~Tian, P.~Isola, A.~Maschinot,
  C.~Liu, and D.~Krishnan, ``Supervised contrastive learning,'' in
  \emph{Advances in Neural Information Processing Systems}, H.~Larochelle,
  M.~Ranzato, R.~Hadsell, M.~F. Balcan, and H.~Lin, Eds., vol.~33.\hskip 1em
  plus 0.5em minus 0.4em\relax Curran Associates, Inc., 2020, pp.
  18\,661--18\,673. [Online]. Available:
  \url{https://proceedings.neurips.cc/paper/2020/hash/d89a66c7c80a29b1bdbab0f2a1a94af8-Abstract.html}
\BIBentrySTDinterwordspacing

\end{thebibliography}

\end{document}